\documentclass[twoside,11pt]{article}

\usepackage{blindtext}
\usepackage{algorithm}
\usepackage{algorithmic}

\usepackage{latexsym, amssymb, amscd, amsthm, amsxtra, amsmath,amsthm, mathtools, mathrsfs}
\usepackage{graphics, graphicx, color}
\usepackage{natbib}
\usepackage{ifpdf}
\usepackage{bm}
\usepackage{xcolor}
\usepackage{xparse}
\usepackage{chngcntr}
\usepackage{kotex}
\usepackage{multirow}     

\usepackage{booktabs}    
\usepackage{caption}     

\usepackage[abbrvbib, preprint]{jmlr2e}

\newcommand{\Bc}{\mathcal{B}}

\newcommand{\Gc}{\mathcal{G}}

\newcommand{\Lc}{\mathcal{L}}
\newcommand{\Mc}{\mathcal{M}}

\newcommand{\Xc}{\mathcal{X}}
\newcommand{\Yc}{\mathcal{Y}}
\newcommand{\Zc}{\mathcal{Z}}

\def \Eb {\mathbb{E}}
\def \Rb {\mathbb{R}}

\def \Pb {\mathbb{P}}

\def\Iv{\mathbf I}

\def\Yv{\mathbf Y}

\def\0v{\mathbf{0}}

\def\1v{\mathbf 1}
\def\0v{\mathbf 0}

\DeclareMathOperator{\thetav}{\bm \theta}

\DeclareMathOperator{\diag}{diag}

\DeclareMathOperator{\vecd}{vecd}

\DeclareMathOperator{\col}{col}
\DeclareMathOperator*{\argmax}{arg\,max}

\usepackage{lastpage}
\jmlrheading{}{2025}{1-\pageref{LastPage}}{10/25}{}{}{Minwoo Kim, Junyong Park, and Sungkyu Jung}

\ShortHeadings{Enhancing Differentially Private Mechanisms via Empirical Bayes}{Kim, Park and Jung}
\firstpageno{1}

\begin{document}

\title{Enhanced Differentially Private Mechanisms via Empirical Bayes}

       
\author{\name Minwoo Kim \email ekfgusdl@snu.ac.kr \\
       \addr Department of Statistics\\
       Seoul National University\\
       Seoul, 08826, South Korea
       \AND
       \name Junyong Park \email junyongpark@snu.ac.kr \\
       \addr Department of Statistics\\
       Seoul National University\\
       Seoul, 08826, South Korea
       \AND
       \name Sungkyu Jung \email sungkyu@snu.ac.kr \\
       \addr Department of Statistics and Institute for Data Innovation in Science\\
       Seoul National University\\
       Seoul, 08826, South Korea}

\editor{My editor}

\maketitle

\begin{abstract}
Differential privacy (DP) has become the gold standard for ensuring the privacy protection of machine learning and statistical algorithms in recent decades. A plethora of algorithms and methods have been developed to enhance the utility of DP algorithms while maintaining the same level of DP. However, these are often overly complex or computationally ineffective. We propose a novel approach focusing on denoising the output of the simple additive Gaussian mechanism by adopting the idea of \textit{empirical Bayes estimation}. We highlight that the empirical Bayes approach can reduce the mean-squared error solely by taking the output of the Gaussian mechanism as input. Our numerical studies show that this simple yet powerful approach can be applied to improve upon various statistical problems, including histogram release, principal component analysis, and linear regression, often outperforming existing private algorithms.
\end{abstract}



\section{Introduction}

In recent decades, data privacy has become a central issue due to the potential risk of sensitive personal information being exposed.
To address this issue, the concept of differential privacy (DP), first proposed by \citet{dwork2006calibrating}, has emerged as a  standard for providing mathematically rigorous privacy guarantees in data analysis and release.
The most widely used and conceptually straightforward approach of ensuring DP is to add  random noise to the target output statistic, an approach known as an \textit{additive mechanism}.
For example, the Gaussian additive mechanism perturbs the output statistic by adding  Gaussian noise.
However, this simple approach often becomes suboptimal, performing well in limited situations, such as when the sample size is large and the output dimension is low. 
To overcome these limitations, numerous studies have proposed problem-specific DP algorithms that improve statistical utility, although they often rely on computationally expensive procedures, sensitive hyperparameter tuning, or unrealistic assumptions such as knowing the true data-generating process. Despite these advances, carefully designed additive mechanisms, such as those used in clipped and perturbed stochastic gradient descent algorithms  \citep{abadi2016deep,chen2020understanding}, remain the workhorses of differential privacy. Improving such additive mechanisms is particularly important because they serve as the foundation for many widely used DP algorithms, and even modest improvements in their utility can translate into significant gains across diverse applications.

In this work, we propose a simple yet powerful approach to enhance the performance of additive differentially private mechanisms through empirical Bayes denoising. Rather than developing new problem-specific algorithms, we focus on post-processing the outputs of existing additive mechanisms by applying empirical Bayes estimators that adaptively recover the underlying signal, or the true unperturbed statistic, from noisy perturbed differentially private outputs. This approach requires no modification of the original privacy mechanism and fully preserves differential privacy through the post-processing property. 
Through extensive experiments, we demonstrate that empirical Bayes post-processing can substantially improve the utility of standard DP mechanisms in diverse settings, including histogram release, principal component analysis, and linear regression.

To illustrate this idea, let $ S = (x_1, \dots, x_n) \in \Xc^n $ be a dataset and 
$ \theta: \Xc^n \to \Rb^p $ be a statistic of interest to be sanitized.
For instance, one might wish to release a sample mean $ \theta(S) = n^{-1}\sum_{i=1}^{n}x_i $, but not directly.
A common approach is to apply the Gaussian additive mechanism:
\begin{equation*}
    \Mc(S) = \theta(S) + \xi, \quad 
    \xi \sim N_p(0, \sigma^2 I_p),
\end{equation*}
where a noise level $ \sigma > 0 $ is calibrated to achieve a specific level of  differential privacy. 
The output can be viewed through the conditional model:
\begin{equation} \label{eq:gauss-additive-model}
    \Mc(S) \:|\: S \sim N_p(\theta(S), \sigma^2 I_p).
\end{equation}
Here, we condition on $S$, since it can be a random sample from an underlying population distribution.

Under model \eqref{eq:gauss-additive-model}, the problem naturally becomes a statistical estimation task: estimating $ \theta(S) $ from the noisy observation $ \Mc(S) $.
Rather than directly using $ \Mc(S) $ as the estimate, one can apply a suitably chosen denoising process to obtain  an improved estimator,
$ \widetilde \Mc(S) $, that reduces the expected squared error
\[
    \Eb \|\widetilde \Mc(S) - \theta(S)\|_2^2 \lesssim \Eb \|\Mc(S) - \theta(S)\|_2^2.
\]
This type of denoising problem has been extensively studied in the statistics and machine learning literature \citep[e.g.,][]{stein1956estimator,efron2012large,efron2024machine}. A classical example is the James-Stein shrinkage estimation \citep{stein1956estimator}, whose idea has been further extended through the empirical Bayes framework \citep{casella1985introduction,brown2009nonparametric,Johnstone05eb}. 

In the seminal work of \citet{balle2018improving}, the authors exploited the post-processing property of differential privacy and applied the classical \textit{James–Stein} shrinkage as a denoising procedure. The key idea of the James-Stein estimator arises from a Bayesian perspective: assuming a mean-zero Gaussian prior for $\theta(S)$, the posterior mean of $\theta(S)$ serves as the denoised output, while the unknown prior variance is estimated from the noisy observation, i.e., $\Mc(S)$. 
This approach improves the utility of the released statistic over the naive output, without any additional privacy loss, thanks to the post-processing property \citep{balle2018improving,kim2025differentially}. However, the effectiveness of the James-Stein denoising is guaranteed only under the restrictive assumptions of Gaussian additive noise and a Gaussian prior distribution. 

To overcome these limitations, we turn to empirical Bayes methods, which provide a more flexible and adaptive framework for denoising when the underlying structure or distribution of  $ \theta(S) $ is unknown. 
The empirical Bayes approach generalizes the James–Stein shrinkage by estimating the prior distribution directly from the noisy data $\Mc(S)$, allowing it to adapt beyond the restrictive Gaussian assumptions.

To operationalize the empirical Bayes post-processing within the differential privacy framework, we adopt two complementary approaches for estimating the prior distribution of the target statistic. 
The first is the nonparametric maximum likelihood estimator (NPMLE) \citep{jiang2009general,koenker2017rebayes}, which provides a fully data-driven, assumption-free estimation of the prior.
The second is the smoothed adaptive shrinkage (SMASH) method \citep{xing2021flexible}, which leverages structural smoothness or sparsity to achieve efficient and adaptive shrinkage in high-dimensional settings. Both approaches are tuning-free and extend naturally beyond the Gaussian setting, making our framework broadly applicable to a wide range of differentially private mechanisms and statistical tasks.

Our main contributions are summarized as follows.


\textbf{A Novel Empirical Bayes Framework}:
We develop a framework of an empirical Bayes mechanism by applying empirical Bayes estimators to the output of differentially private additive mechanisms, thereby enhancing the utility of the simple additive mechanism and improving the utility-privacy trade-off.
To the best of our knowledge, this is the first work that systematically applies empirical Bayes estimators to various statistical estimation problems under the constraint of differential privacy.

\textbf{Flexibility and Adaptivity}:
The empirical Bayes framework is highly flexible on the distributional or structural knowledge of $\theta(S)$.
A fully nonparametric approach that does not require any assumptions on $ \theta(S) $ is possible by the NPMLE approach (Section \ref{sec:NPMLE}).
On the other hand, a model-based empirical Bayes method such as SMASH (Section \ref{sec:SMASH}) can also significantly enhance the efficiency of denoising if the assumed model, such as structural sparsity on $\theta(S)$, is nearly correct. 
Additionally, it is a fully adaptive approach that relies solely on the observed data.
    
\textbf{Broad Applicability and Effectiveness}: 
Specifically, we apply this approach to practical statistical problems such as histogram release, principal component analysis, and linear regression. 
As a consequence, it dramatically enhances the utility of noise additive mechanisms in all such statistical problems.
We confirm that empirical Bayes approaches improve upon recently proposed DP methods in enhancing the utility, especially in high dimensions and under high privacy guarantees. 
    
Although we adopt the  recently proposed  privacy notion of Gaussian differential privacy \citep{dong2022gaussianDP}, our approach can also be applied to other definitions, such as $\varepsilon$-DP \citep{dwork2006calibrating}.
In the appendix, we illustrate how our empirical Bayes framework can be applied to the Laplace mechanism, which guarantees $\varepsilon$-DP.
As a denoising post-processing, empirical Bayes could be applied to broader DP mechanisms beyond the additive format.

\subsection{Related Work}
Utilizing the denoising process as post-processing to enhance the utility of DP mechanisms for the statistical estimation problems has been frequently adopted in DP research.
For example, this idea has been applied in contingency table release \citep{barak2007privacy}, network analysis \citep{hay2009accurate}, graph generation \citep{karwa2016inference, bernstein2017differentially}, and deep learning \citep{nasr2020improving, wang2021dplis, cheng2022improved, yan2023deep}.
Notably, \citet{balle2018improving} proposed using statistical denoising methods, such as the James-Stein estimator, which is a fully adaptive approach that does not rely on any prior knowledge.

While we consider the central model of differential privacy, the iterative Bayesian update (IBU) has been studied in the context of the local model of differential privacy (e.g., \cite{elsalamouny2020generalized} or \cite{Arcolezi2023iterative}). 
It is worth noting that these IBU approaches differ from the empirical Bayes scheme in that IBU aims to estimate the underlying distribution itself. 
In contrast, empirical Bayes aims to estimate the target mean parameters from noisy observations when we do not know the prior distribution.

\section{Preliminary on Differential Privacy}

In this section, we briefly review Gaussian differential privacy (GDP)  and related concepts proposed by \citet{dong2022gaussianDP}. 
GDP is a single-parameter privacy notion that compares the worst-case trade-off function of a given mechanism to the trade-off function between mean-shifted Gaussian distributions.

Let $S = (x_1, \dots, x_n) \in \Xc^n$ be a dataset, where $\Xc$ denotes a data space.
We say a dataset $S' = (y_1, \dots, y_n)$ is neighboring to $S$ if $x_i = y_i$ for all but one index $\iota$ such that $x_\iota \neq y_\iota$.
We denote this neighboring relation as $S \sim S'$.

Let $P$ and $Q$ be probability distributions. 
Consider a statistical hypothesis testing problem to determine whether 
the observed sample $ X $ comes from $P$ or $Q$:
\begin{equation} \label{eq:test}
H_0: X \sim P \quad \text{v.s.} \quad H_1: X \sim Q.
\end{equation}
Trade-off function is defined as the minimum type II error rate achievable at a given type I error rate for the test \eqref{eq:test}.
Let $\phi$ be a test function which takes $X$ and outputs value in $[0, 1].$ 
In other words, we reject $H_0$ with probability $\phi(X)$.
Then the type I error $\alpha_\phi$ and the type II error $\beta_\phi$ are defined as 
\begin{equation*}
    \alpha_\phi = \Eb_{X \sim P}[\phi(X)] \quad \mbox{and} \quad
    \beta_\phi = \Eb_{X \sim Q}[(1 - \phi(X))].
\end{equation*}

\begin{definition}
Consider the test \eqref{eq:test}.
The trade-off function between $P$ and $Q$, $T(P, Q): [0, 1] \to [0, 1]$ is defined as 
\[
T(P, Q)(\alpha) = \inf \left\{ \beta_\phi \:|\: \alpha_\phi \le \alpha  \right\},
\]
where the infimum is taken over all test function $ \phi $.
\end{definition}

Intuitively, the trade-off function indicates how difficult it is to distinguish between the given distributions in terms of hypothesis testing.

Next, we introduce the notion of Gaussian differential privacy.
Let $ \mu > 0 $. Consider a test between Gaussian distributions:
\[
    H_0: X \sim N(0, 1) \quad \text{v.s.} \quad H_1: X \sim N(\mu, 1).
\]
As a direct consequence of the Neyman-Pearson lemma, it can be shown that 
\[
T(N(0, 1), N(\mu, 1))(\alpha) =  \Phi(\Phi^{-1}(1-\alpha) - \mu),
\]
where $\Phi$ is the cumulative distribution function of $N(0, 1)$.
Then, the Gaussian differential privacy is defined as follows.
\begin{definition}
    Let $ \mu > 0 $ be a privacy parameter.
    Then a randomized mechanism $ \Mc $ satisfies 
    $ \mu $-\textit{Gaussian differential privacy} (GDP) if 
    for each $ \alpha \in [0, 1] $, it holds that 
    \[
        \inf_{S \sim S'} T(\Mc(S), \Mc(S'))(\alpha) \ge T(N(0, 1), N(\mu, 1))(\alpha),
    \]
    where the infimum takes over all possible pair of neighboring datasets.
\end{definition}

Suppose that $\Mc$ is $\mu$-GDP.
Then for any neighboring datasets $S \sim S'$, distinguishing 
the outputs of $\Mc(S)$ and $\Mc(S')$ by a hypothesis test is 
at least as difficult as distinguishing two Gaussian distributions, 
$N(0, 1)$ and $N(\mu, 1)$.
When the privacy parameter $ \mu $ has a low value, it gives 
high privacy protection because $ N(0, 1) $ and $ N(\mu, 1) $ are 
more likely to indistinguishable.
Conversely, a high $\mu$ value offers low privacy protection.

Consider a multivariate statistic $ \theta(S) \in \Rb^p $ for a dataset
$S$ which we want to sanitize and to make satisfy GDP.
The $ \ell_2 $-sensitivity of $ \theta $ is defined as 
\[
    \Delta_2(\theta) := \sup_{S \sim S'} \|\theta(S)-\theta(S')\|_2.
\]
To convert $\theta(S)$ to satisfy $\mu$-GDP, one of the simplest
approach is adding a Gaussian noise with variance scaled to $\Delta_2(\theta)$ to $\theta(S)$.
\begin{proposition}[\citet{dong2022gaussianDP}] \label{prop-Gmech}
    Let Gaussian additive mechanism 
    $ \Mc_G(S; \sigma) = \theta(S) + \xi $, where $ \xi \sim N_p(0, \sigma^2 I_p) $.
    Then $ \Mc_G $ satisfies $ \mu $-GDP if $ \sigma \ge \Delta_2 / \mu $.
    In the rest of the paper, we just write  $ \Mc_G $ by fixing $ \sigma = \Delta_2 / \mu $.
\end{proposition}

We wrap up this preliminary section with important features of GDP: post-processing property and composition of multiple mechanisms.
These enable designing complex private algorithms from 
simpler ones.

\begin{proposition} \label{prop:post-processing}
    Consider a $ \mu $-GDP mechanism $ \Mc: \Xc \to \Yc $.
    Let $ \mbox{Proc}: \Yc \to \Zc $ be a (possibly randomized) post-processing 
    function which is independent to data.
    Then, $ \mbox{Proc} \circ \Mc: \Xc \to \Zc $ also satisfies $ \mu $-GDP.
\end{proposition}

\begin{proposition}
    Suppose that $ \Mc_i $ is $ \mu_i $-GDP for each $ i = 1, \dots, k $.
    Then their composition mechanism $ \Mc = (\Mc_1, \dots, \Mc_k) $ is $ \mu $-GDP 
    where $\mu = \sqrt{\mu_1^2 + \dots \mu_k^2}.$ 
    This also holds in an adaptive setting where $ \Mc_i $ may takes the output of 
    $ \Mc_1, \dots, \Mc_{i-1} $ as inputs.
\end{proposition}

\section{Empirical Bayes Estimator} \label{sec-EB}

In this section, we review the methods of empirical Bayes. 
Empirical Bayes approaches can denoise the output of differentially private mechanisms and enhance the utility while maintaining privacy level in various statistical problems.

Consider the following data model for a mechanism output $\Mc(S) = \Yv = (Y_1, \dots, Y_p)^\top$:
\begin{equation} \label{eq:eb-datamodel}
    Y_i \:|\: \theta_i \overset{ind.}{\sim} p(y \:|\: \theta_i) ~\text{ and }~
    \theta_1, \dots, \theta_p \overset{i.i.d.}{\sim} G^*.
\end{equation}
Here, $G^*$ is an unknown prior distribution of the parameters $\theta_i$, but  $p(\cdot \:|\: \theta) $  is typically known. For example, when the Gaussian additive mechanism  \eqref{eq:gauss-additive-model} is used, the distribution $p(\cdot \:|\: \theta) $ is exactly known. 
We only observe $\Yv = (Y_1, \dots, Y_p)^\top$, while
$ \thetav = (\theta_1, \dots, \theta_p) $ are unobserved.
In this stage, the goal is estimating $ \thetav$ only from $ \Yv $.

Under the Bayesian perspective, it is well-known that the Bayes estimator
\[
    \hat\theta_i^* := \Eb_{G^*} [\theta \:|\: Y_i],
\]
where $ \theta \sim G^*$, 
minimizes the Bayes risk of the squared error $ (Y_i - \theta)^2 $.
The main idea of \textit{empirical Bayes} is to substitute the unknown prior $ G^* $ with $ \widehat G_p $, which is estimated from the observed data $\Yv$.
As the result, the \textit{empirical Bayes estimator} of each $\theta_i$ is defined as
\begin{equation} \label{eq:EBdef}
    \hat\theta_i \equiv \hat \theta_i(Y_1, \dots, Y_p) := \Eb_{\widehat G_p} [\theta \:|\: Y_i].
\end{equation}
When the empirical prior distribution $ \widehat G_p $ well approximates $ G^* $, we expect that $ \hat \theta_i \approx \hat \theta_i^*.$

\subsection{The James-Stein Estimator}

The James-Stein estimator is a special case of empirical Bayes estimator, obtained under the Gaussian perturbation $p(\cdot | \theta)$ and prior distribution $G^*$. More specifically,   
%
%
assume a Gaussian prior 
$ \theta(S) \sim N_p(0, \omega^2 I_p) ,$ for some $ \omega > 0 $, for the Gaussian additive mechanism model \eqref{eq:gauss-additive-model}. 
Then it is well known that the Bayes estimator becomes 
\begin{equation*}
    \tilde \theta_B(S) = \left(1 - \frac{\sigma^2}{\sigma^2 + \omega^2}  \right) \Mc(S).
\end{equation*}
In general, $ \omega $ is unknown but it can be estimated from the 
\textit{observation} $ \Mc(S) $.
Since $ \Mc(S) \sim N_p(0, (\sigma^2 + \omega^2)I_p) $, 
it holds that $ (\sigma^2 + w^2)^{-1} \|\Mc(S)\|_2^2 \sim \chi^2_p $. 
From this, for $ p \ge 3 $,
\begin{equation*}
    \Eb\left[ \frac{1}{\|\Mc(S)\|_2^2}\right]
    = \frac{1}{\sigma^2 + \omega^2 } \cdot \frac{1}{p-2}.
\end{equation*}
Plugging-in this to the Bayes estimator $ \tilde \theta_B(S) $, we obtain the James-Stein estimator
\begin{equation*}
    \hat \theta_{JS}(S) = 
    \left(1 - \frac{(p-2)\sigma^2}{\|\Mc(S)\|_2^2} \right) \Mc(S).
\end{equation*}
Importantly, it holds that 
$\Eb \|\hat \theta_{JS}(S) - \theta(S)\|_2^2
    < \Eb \|\Mc(S) - \theta(S)\|_2^2. $
In addition, $ \hat \theta_{JS} $ guarantees the same level of differential privacy as $ \Mc(S) $ due to the post-processing property.

The main idea of the above example lies in estimating the unknown prior distribution (e.g., $ \omega^2 $) solely from the observation (e.g., $ \Mc(S) $), and substituting it into the Bayes estimator. 
The \textit{Empirical Bayes} approaches adapt well beyond Gaussian prior distributions, making the James-Stein estimator as a special case. 
%
%
Several methods of empirical Bayes outperform when the model is mis-specified for the James-Stein setting, i.e., $\theta(S) \sim G^*$ does not follow Gaussian. Two empirical Bayes methods for estimating $G^*$ soley from the mechanism output $\Yv$ are discussed next.

\subsection{Estimating $G^*$ by NPMLE} \label{sec:NPMLE}
One popular approach for estimating $G^*$ solely from the observed data $\Yv$ is the \textit{nonparametric maximum likelihood estimator} (NPMLE).

Under the model \eqref{eq:eb-datamodel}, the marginal likelihood 
can be written as
\begin{equation}
    \Lc(Y_1, \dots, Y_p; G^*) = \prod_{i=1}^p f_{G^*}(Y_i),
\end{equation}
where $ f_{G^*}(y) = \int p(y|\theta) dG^*(\theta)$ is the marginal density function of $Y_i$'s.
Then the NPMLE $ \widehat G_p $ of $G^*$ is defined as 
\begin{align}
    \widehat G_p &:= \argmax_{G \in \Gc} \Lc(Y_1, \dots, Y_p; G) \nonumber \\
                 &= \argmax_{G \in \Gc} \prod_{i=1}^p f_{G}(Y_i), \label{eq:npmle}
\end{align}
where $ \Gc $ is a class of prior distributions.
Once the NPMLE $\widehat G_p$ is obtained, the empirical Bayes estimator \eqref{eq:EBdef} can be obtained as 
\[
    \hat \theta_i = \Eb_{\widehat G_p}[\theta | X_i]
    = \frac{\int \theta p(Y_i \:|\: \theta) d\widehat G_p(\theta)}{\int p(Y_i \:|\: \theta) d\widehat G_p(\theta)}.
\]

For the largest possible class $\Gc$, surprisingly, a discrete distribution of the form $\widehat G_p = \sum_{i=1}^m w_i \delta_{a_i} $ is known to be a maximizer of \eqref{eq:npmle} \citep{lindsay1995mixture, ignatiadis2025empirical}.
Based on this fact, \citet{jiang2009general} proposed to use expectation-maximization algorithm to find an approximate NPMLE with restricted $\Gc$ consists of distributions finitely supported on a pre-determined grid over $ [\min_i Y_i, \max_i Y_i] $; see the supplementary for a detail.
%
An efficient solver for this process 
is implemented in R package \texttt{REBayes} \citep{koenker2017rebayes}.
We remark that the above procedure for obtaining NPMLE is valid for random perturbation of any kind, as well as for Gaussian noises.

\subsection{Smoothed Adaptive Shrinkage EB} \label{sec:SMASH}
Here, we suppose that $ p(\cdot \:|\: \theta_i) $ follows a
Gaussian distribution: $\Yv | \thetav \sim N_p(\thetav, \sigma^2 I_p).$ 
Note that this is a valid setting for the Gaussian additive mechanism.
Under the Gaussian assumption, we utilize the empirical Bayes approach called \textit{smoothing by adaptive shrinkage}, proposed by  \citet{xing2021flexible}.

Let $W$ be $p \times p$ an orthogonal matrix of discrete wavelet transform for $\Yv$. 
By multiplying $W$, we have 
\[
    W \Yv \;|\; W\thetav \sim N_p(W\thetav, \sigma^2 I_p),
\]
since $WW^\top = I_p$.
Denote $ \widetilde \Yv =  W \Yv $ and $ \widetilde \thetav = W \thetav $.
When $\thetav = (\theta_1, \dots, \theta_p)$ is \textit{spatially structured} (i.e., $ \theta_i $ and $ \theta_j $ have similar value for small $ |i - j| $), many elements of $ \widetilde \thetav $ become close to zero, making it approximately sparse \citep{donoho1995adapting, mallat1999wavelet, xing2021flexible}.
Aiming for spatially structured $\theta$, \citet{xing2021flexible} proposed to fit empirical Bayes estimator on 
\begin{equation*}
    \widetilde \Yv \;|\; \widetilde \thetav \sim N_p(\widetilde \thetav, \sigma^2 I_d),
    \quad \tilde \theta_i \overset{i.i.d.}{\sim} G,
\end{equation*}
with the restricted family of the prior distribution of the form
$G = \sum_{k=0}^{K}\pi_k N(0, \omega_k^2),$
including zero-centered scale mixtures of Gaussian distributions. 
This choice of prior is well-suited for shrinking the noisy coefficients in $\widetilde{\Yv}$ towards zero, effectively leveraging the sparsity of $\widetilde{\thetav}$.
Let $ \widehat{\thetav}^{EB} $ be the resulting empirical Bayes estimate of $\widetilde \thetav$ based on $\widetilde \Yv$.
Final estimator of $ \theta $ then can be obtained by 
\[
    \widehat \thetav^{SMASH} := W^{-1}\widehat{\thetav}^{EB} = W^\top \widehat{\thetav}^{EB}.
\]
Here, SMASH is an abbreviation of ``SMoothing by Adaptive SHrinkage.''
The estimating procedure of SMASH is publicly available through R package \texttt{smashr}.

\begin{remark} 
The SMASH estimator does not assume any specific parametric form for the prior distribution. In practice, it approximates the prior using a Gaussian mixture, which is flexible enough to represent a wide variety of prior shapes. Although the estimator is developed under the Gaussian additive mechanism, its computation depends only on the observed data $\mathbf{Y}$ and the known noise variance $\sigma^2$ of the additive perturbation.
This property allows SMASH (and likewise NPMLE) to be interpreted as a general-purpose denoising procedure for additive Gaussian mechanisms, and potentially for other noise-adding mechanisms such as the Laplace mechanism.
\end{remark}

\section{Proposed Methods}
Let $S = (x_1, \dots, x_n)$ be a dataset at hand.
We are interested in sanitizing a statistic of interest $\theta: (\Rb^d)^n \to \Rb^p$ for public release.
Consider an additive Gaussian mechanism:
\begin{equation*}
    \Mc_G(S) = \theta(S) + N_p\left(0, \:\frac{\Delta_2^2}{\mu^2}\Iv_p\right),
\end{equation*}
A simple yet important observation is that the following conditional distribution relationship
\[
\Mc_G(S) \:|\: S \overset{d}{=} \Mc_G(S) \:|\: \theta(S) \sim   N_p\left(\theta(S),\:\frac{\Delta_2^2}{\mu^2}\Iv_p\right)
\]
matches  the model \eqref{eq:eb-datamodel} by regarding $\Yv \leftarrow \Mc_G(S)$ and $\thetav \leftarrow \theta(S) $, since $p(\cdot\:|\:\theta)$ is the density function of Gaussian distribution with known $ \sigma = \Delta_2 / \mu $, while the prior distribution of $\theta(S)$ is unknown.
By treating the post-processing of $\Mc_G(S)$ as estimating $\theta(S)$ based on the observation $\Mc_G(S)$, we apply the empirical Bayes estimators considered in the previous section. 
This can be viewed as a denoising procedure that adaptively reduces the variability introduced by the additive noise, and thus is expected to enhance the utility compared to directly releasing the output of $\Mc_G$.
Since this process is a post-processing step, the privacy guarantee of $\Mc_G(S)$ is not compromised (for e.g., see Proposition \ref{prop:post-processing}).
We formally define the empirical Bayes post-processing mechanisms as follows.

\begin{definition}
    Let $\mbox{Proc}_{EB}(\eta; \sigma)$ be an empirical Bayes estimator of $\eta \in \Rb^p$, where the observation is sampled from $ N_p(\eta, \sigma^2 \Iv_p)$. 
    The \textit{empirical Bayes} (EB) mechanism for a statistic $\theta(S) \in \Rb^p$ with scale parameter $\sigma > 0$ is defined as
    \[
    \Mc_{EB}(S; \sigma) := 
    \mbox{Proc}_{EB}(\Mc_G(S; \sigma); \sigma).
    \]
\end{definition}
Under the DP setting, $ \sigma $ is generally known since it is determined a priori by users to satisfy a pre-specified level of DP. 
Thus, all empirical Bayes approaches introduced in the previous section can be naturally applied to the output from $ \Mc_G(S) $ and it is well-defined.
Also, based on the post-processing property, the EB mechanism satisfies DP.
\begin{theorem}
   Let $\sigma \ge \Delta_2(\theta)/\mu$.
   Then $ \Mc_{EB}(S; \sigma) $ satisfies $ \mu $-GDP.
\end{theorem}
In the rest of the paper, we fix $ \sigma = \Delta_2(\theta)/\mu $, and omit $\sigma$ from the notation of the empirical Bayes mechanism.

\begin{remark}
    The Empirical Bayes post-processing can be applied to other types of additive mechanisms. As an example, the utility of the Laplace mechanism, ensuring $ \varepsilon $-DP, can be enhanced by applying EB-NPMLE; see Appendix \ref{appendix:Lap} for a detailed discussion.
    %
   %
\end{remark}

\section{Applications}
In this section, we illustrate how our empirical Bayes framework can be applied to various statistical tasks.
Differentially private versions of histogram release, principal component analysis, and linear regression are investigated in the following subsections.

\subsection{Private Histogram Release}
Let $ S = (x_1, \dots, x_n) $ be a dataset where each observation $ x_i $ belongs to a data space $\Xc$.
Consider a partition $\Bc = \{B_1, \dots, B_p\}$ of $ \Xc $, i.e., $ \Xc = \cup_{i=1}^p B_i $ and $ B_i \cap B_\iota = \emptyset $ for $ i \neq \iota $.
Let the histogram (or vectorized frequency count table) over the partition $\Bc$ be $\theta^{hist}(S) = (\theta_1(S), \dots, \theta_p(S))$, where
\[
    \theta_i(S) = |\{j : x_j \in B_i\}|.
\]
For neighboring datasets $S \sim S'$, it holds that
$ \theta^{hist}(S) - \theta^{hist}(S') = e_i - e_j $ for some $ i, j \in \{1, \dots, p\} ,$
where $ [e_1, \dots, e_p] = I_p $.
Thus, the $\ell_2$-sensitivity of the histogram is $\Delta_2(\theta^{hist}) = \sqrt{2}$.
By applying the Gaussian additive mechanism, we can obtain a $ \mu $-GDP histogram:
\[
    \Mc_G^{hist}(S) := \theta^{hist}(S) + \xi, 
    \quad \xi \sim N_p\left(0, \frac{2}{\mu^2}I_p\right).
\]
Proposition \ref{prop-Gmech} implies that $ \Mc_G^{hist} $ satisfies $ \mu $-GDP,
and we can apply EB mechanisms as $ \mbox{Proc}_{EB}(\Mc_G^{hist}) $.

\subsubsection{Related Work}
The private release of histograms has been studied from the early stage of development of DP researches \citep{dwork2006calibrating, wasserman2010statistical}. 
While various methods exist, our work is most related to those that improve utility via post-processing.
For example, a rank-reduced James-Stein mechanism was recently
proposed by \citet{kim2025differentially} to denoise histogram under $ \mu $-GDP, leveraging the rank-deficient structure of the histogram and James-Stein estimator as a denoising.

\subsection{Private Principal Component Analysis}
In this subsection, we apply the empirical Bayes mechanism to the recently proposed differentially private PCA method of \cite{kim2025pca}. 

Let $\mbox{Sym}(d)$ denote the set of $d \times d$ symmetric matrices.
Following \cite{schwartzman2016lognormal}, define vectorization operator $\vecd: \mbox{Sym}(d) \to \Rb^{d(d+1)/2}$ as
\[
A \mapsto (\diag(A)^{\top}, \sqrt{2} \mbox{offdiag}(A)^{\top})^{\top}.
\]
One key advantage of $ \vecd $ is that it preserves the Frobenius norm, i.e., $\|A\|_F = \|\vecd(A)\|_2$ for any $A \in \mbox{Sym}(d)$.

Let $S = (x_1, \dots, x_n) \in (\Rb^d)^n$ be a dataset.
\citet{kim2025pca} proposed a robust and private PCA mechanism by applying the matrix additive Gaussian mechanism to the Kendall's tau matrix \citep{han2018eca} $ \widehat{K}(S) $ defined as
\[
    \widehat K(S)
    = \frac{2}{n(n-1)} \sum_{i < j} \frac{(x_j - x_i)(x_j - x_i)^\top}{\|x_j - x_i\|_2^2}.
\]
When the $x_i$'s are drawn from an elliptical distribution, the eigenvectors of $\widehat K$ coincide with those of the population covariance matrix.
Here, the family of elliptical distribution includes multivariate Gaussian and heavy-tailed distributions such as the multivariate $t$.\footnote{See \citet{fang2018symmetric} for a general introduction on elliptically distributed data.}
Therefore, extracting principal components (PCs) from $\widehat K$ constitutes a valid PCA procedure.

Due to the self-normalization process, it can be verified that $\Delta_2(\vecd(\widehat K)) \le 4/n.$ 
Then for $ \sigma \ge 4/(n\mu) $, the (matrix-variate) Gaussian additive mechanism
\[
    \widetilde K(S; \sigma)
    = \widehat K(S) + \vecd^{-1}(\xi), \quad \xi \sim N_{q}(0, \sigma^2 I_{q})
\]
satisfies $ \mu $-GDP, where $q = d(d+1)/2$.
We set $ \sigma = 4/(n\mu) $ for $ \widetilde K $. 
Once $\widetilde{K}(S) = \widetilde K(S; \sigma)$ is released, its top eigenvectors can be used as the $\mu$-GDP principal components.

Since Gaussian noise is added to ensure privacy, the empirical Bayes post-processing mechanisms can be directly applied to $\widetilde K$.
Private and denoised PCs then can be obtained by the eigen-decomposition of $\mbox{Proc}_{EB}(\widetilde K) $.

\subsubsection*{Related Work}
In past decade, private PCA has been extensively studied. Early approaches used additive mechanism on the sample covariance matrix \citep{chaudhuri2013near, dwork2014analyze}. 
Recent works achieve minimax optimality under Gaussian assumptions but are either computationally expensive, limited to the first PC, or rely on relaxed privacy notions \citep{liu2022dp, asi23robustdp, cai2024optimal}. 
Iterative methods like the noisy power method \citep{hardt14noisypower} have been explored as well.
Notably, \citet{maunu2022stochastic} developed a noisy geodesic gradient descent (NSGGD) on the Grassmanian manifold for private and robust PC estimation.

\subsection{Private Linear Regression}
We consider linear regression as the final application and apply an EB mechanism to the adaptive sufficient statistics perturbation (AdaSSP) proposed by \citet{wang2018revisiting}, a widely used method for DP linear regression.

Let $X \in \mathbb{R}^{n \times {(d+1)}}$ be a design matrix where the first column consists of $1_n = (1, \dots, 1)^\top$ and $Y \in \mathbb{R}^n$ denote the response vector.
The goal is privately estimating $\beta \in \Rb^{d+1}$ under the linear model 
$$
Y = X\beta + e, \quad \mbox{where } ~~ e \sim (0, \sigma^2 I).
$$
Recall that the non-private OLS estimator is given as
\[
\hat \beta^{ols} = (X^\top X)^{-1}X^\top Y,
\]
and the ridge estimator has the form of 
\[
\hat \beta_{\lambda} = (X^\top X + \lambda I)^{-1}X^\top Y,
\]
where $\lambda > 0$ is a tuning parameter.
AdaSSP consists of two main parts: (i) adding noise to each sufficient statistics $X^\top X$ and $X^\top Y$, and (ii) privately selecting the parameter $\lambda$ to minimize the expected empirical risk.
We consider Gaussian mechanisms to perturb $X^\top X$ and $X^\top Y$ as follows:
\begin{align*}
    \widehat{X^\top X} &= X^\top X +  \vecd^{-1}(\xi_1), 
    \quad \xi_1 \sim N(0, \sigma_1^2 I_{d})\\
    \widehat{X^\top Y} &= X^\top Y + \xi_2, \quad \xi_2 \sim N(0, \sigma_2^2 I_d),
\end{align*}
where $\sigma_1 = 2\|\Xc\|\sqrt{1+\|\Xc\|^2} / (\mu / \sqrt{3})$ and
$\sigma_2 = 2\|\Yc\|\sqrt{1+\|\Xc\|^2} / (\mu / \sqrt{3})$.
Here, $\sigma_1$ and $\sigma_2$ need to be scaled to make each of $\widehat{X^\top X}$ and $\widehat{X^\top Y}$ satisfy $ \mu / \sqrt{3} $-GDP.\footnote{To ensure $\sigma_1, \sigma_2 < \infty$, it needs that the space of data $\mathcal{X}$ and $\mathcal{Y}$ are bounded. 
If it is needed, we clip the dataset to make $\mathcal{X}$ and $\mathcal{Y}$ finite.}
Then, we propose to apply EB mechanisms to $\widehat{X^\top X}$ and
$\widehat{X^\top Y}$.
Denote the resulting outputs as $\widehat{X^\top X}_{EB}$ and
$\widehat{X^\top Y}_{EB}$.
The tuning parameter $\hat \lambda$ is set as same as the original AdaSSP mechanism using the privacy budget of $\mu / \sqrt{3}$.
Then the final EB-AdaSSP mechanism is given as
\[
    \hat \beta^{EB} = (\widehat{X^\top X}_{EB} + \hat \lambda)^{-1}\widehat{X^\top Y}_{EB}.
\]

\subsubsection*{Related Work}
Several approaches have been proposed for DP linear regression. 
The Sufficient Statistic Perturbation (SSP) \citep{zhang12functional}
adds noise directly to $X^\top X$ and $X^\top Y$ prior to computing the OLS estimator. 
As we previously mentioned, AdaSSP \citep{wang2018revisiting} extends SSP by incorporating ridge regression and adaptively selecting the ridge parameter. 
\citet{Amin2023EasyDP} introduced the Tukey exponential mechanism, which combines subsample-based OLS with private selection via Tukey depth, requiring large sample sizes ($n \ge 1000d$) but without assuming bounded data. 
Gradient-based methods have also been explored: \citet{brown2024private} proposed noisy gradient descent using per-sample clipping, and \citet{tang2024improved} combined gradient boosting with AdaSSP. 

\section{Numerical Studies}
In this section, we numerically explore how empirical Bayes mechanism enhances  the statistical utility compared to the existing DP algorithms in the cases of releasing histogram, PCA and linear regression.
Throughout our numerical studies, we implement two empirical Bayes (EB) mechanisms as explained in Section \ref{sec-EB}. 
For the EB-NPMLE, we use the R package \texttt{REBayes} \citep{koenker2017rebayes}, and for the EB-SMASH, we use the R package \texttt{smashr} \citep{xing2021flexible}.


\begin{figure}[]
	\centering
	\includegraphics[width=.9\linewidth]{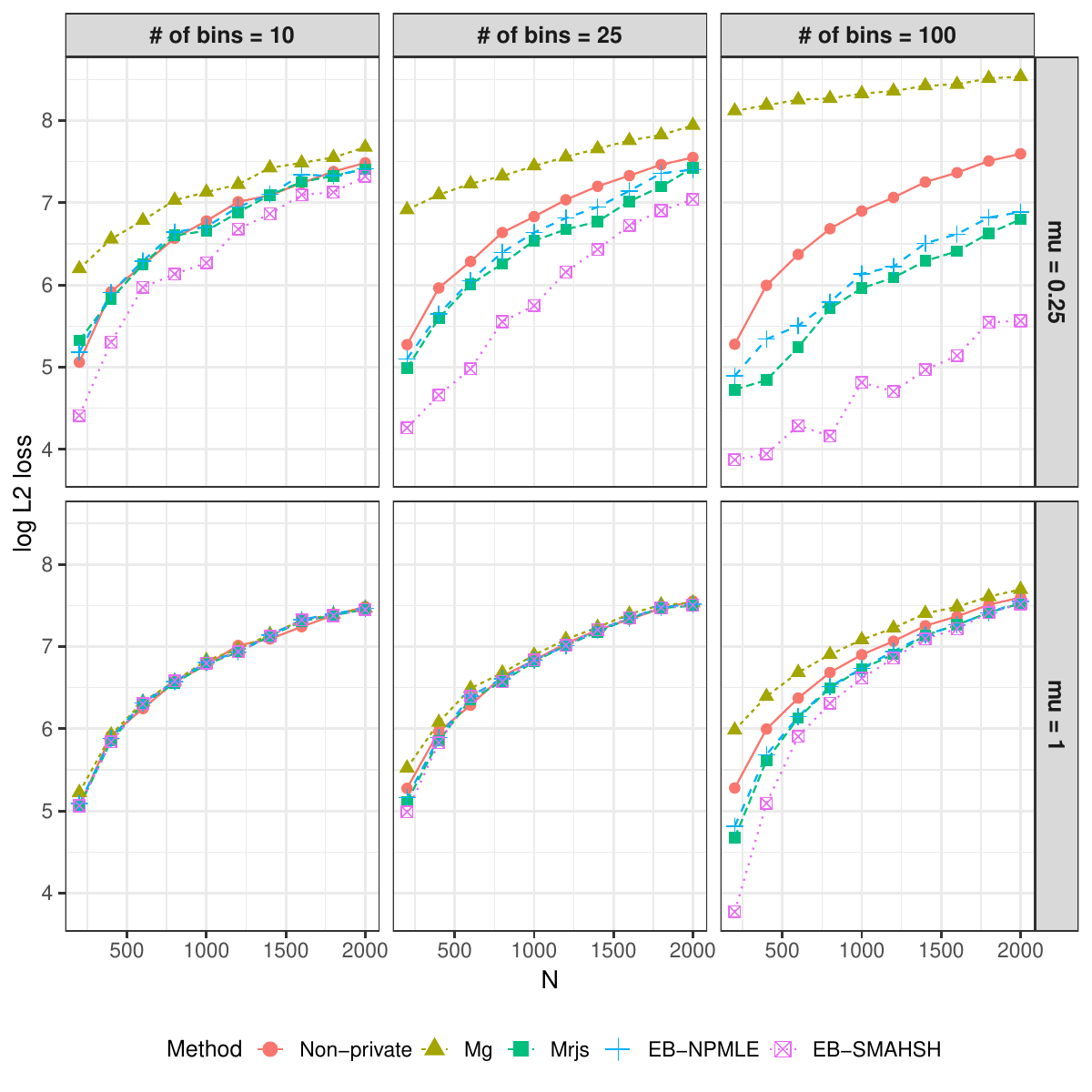}
    \caption{Average population $L_2$-costs of private histogram release under Model I.}
    \label{fig:hist-M1}
\end{figure}

\begin{figure}[]
	\centering
	\includegraphics[width=.9\linewidth]{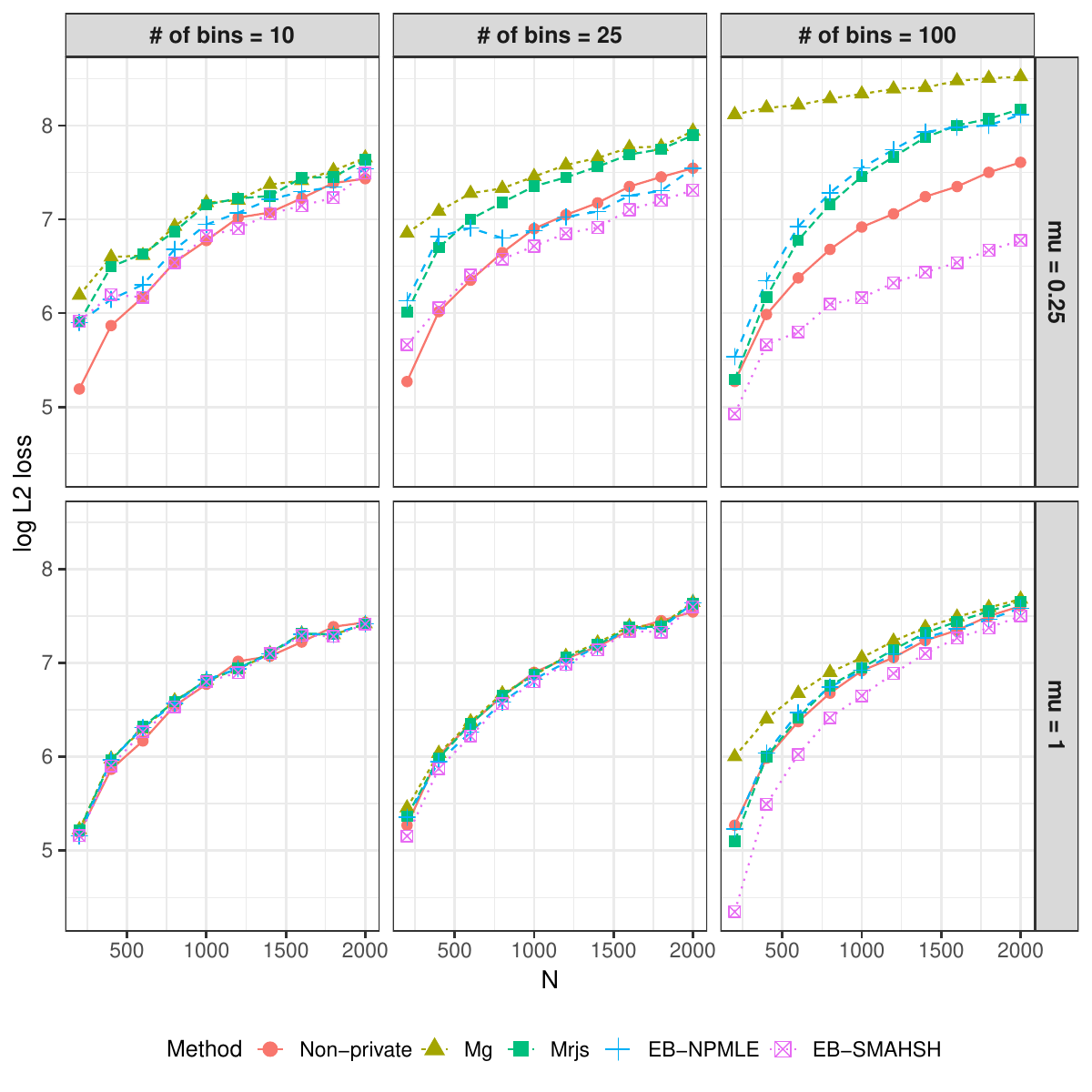}
    \caption{Average population $L_2$-costs of private histogram release under Model II.}
    \label{fig:hist-M2}
\end{figure}

\subsection{Private Histogram Release}
In this section, the utility in terms of $L_2$-costs of the proposed empirical Bayes post-processing mechanisms are demonstrated with numerical simulations. 
We generate true histogram counts $\thetav(S)$ from a multinomial distribution $\mbox{Multinomial}(N, \pi)$, where $\pi = (\pi_1, \dots, \pi_p)$ such that $\sum_{i=1}^p \pi_i = 1$. 
Two models for $\pi$ are considered as follows:
\begin{itemize}
    \item Model I (constant): $\pi_1 = \pi_2 = \dots = \pi_p = \frac{1}{p}$,
    \item Model II (two-peaks): $ \pi_j =  \frac{1}{4p} \text{ for } j = 1, \dots, \frac{p}{4}  \text{ and } j = \frac{3p}{4} + 1, \dots, p,$ and 
$\pi_j = \frac{1}{12p} \text{ for } j = \frac{p}{4} + 1, \dots, \frac{3p}{4}.$
\end{itemize}
For parameters, we consider $p \in \{10, 25, 100\}$, $ \mu \in \{0.25, 1\} $ and $N \in \{200, 400, \dots, 2000\}.$
For each combination of model and $ (N, p, \mu) $, we executed various
$ \mu $-GDP mechanisms $ \Mc(S) $ to sanitize $ \theta(S) $, and recorded
the population $ L_2 $-cost $ \|\Mc - N\pi\|_2^2 $. This is repeated for 100 times, and the 
averaged $L_2$-costs are reported in Figures \ref{fig:hist-M1} and \ref{fig:hist-M2}.
For the competing models, we consider the naive Gaussian mechanism $\Mc_G$ and the rank-deficient James-Stein mechanism $\Mc_{rJS}$ introduced at \citet{kim2025differentially}. 
We note that $\Mc_{rJS}$ outperforms James-Stein mechanism considered in \citet{balle2018improving} both theoretically and empirically \citep{kim2025differentially}.
The performance of the non-private histogram $\thetav(S)$ is also plotted for reference.

%

In both figures, EB-SMASH outperforms all the other methods.
We note that EB methods are always better than just using the Gaussian mechanism, $ \Mc_G $.
While the rank-deficient James-Stein mechanism ($\Mc_{rJS}$) also reduces the $L_2$-cost compared to the basic Gaussian mechanism, our proposed EB methods outperform $\Mc_{rJS}$.
A key finding is that the performance gains from our EB methods are most pronounced in high-dimensional ($p=100$) and high-privacy ($\mu=0.25$) settings.
Notably, in these scenarios, EB-SMASH can even outperform the non-private histogram.
Conversely, when the dimension is low (e.g., $p=10$) and the privacy
guarantee is weaker ($\mu=1$), the performance gap between the
various private mechanisms narrows, and the denoising methods do not significantly improve the Gaussian mechanism.

\subsection{Private Principal Component Analysis}
\begin{figure}[]
	\centering
	\includegraphics[width=.95\linewidth]{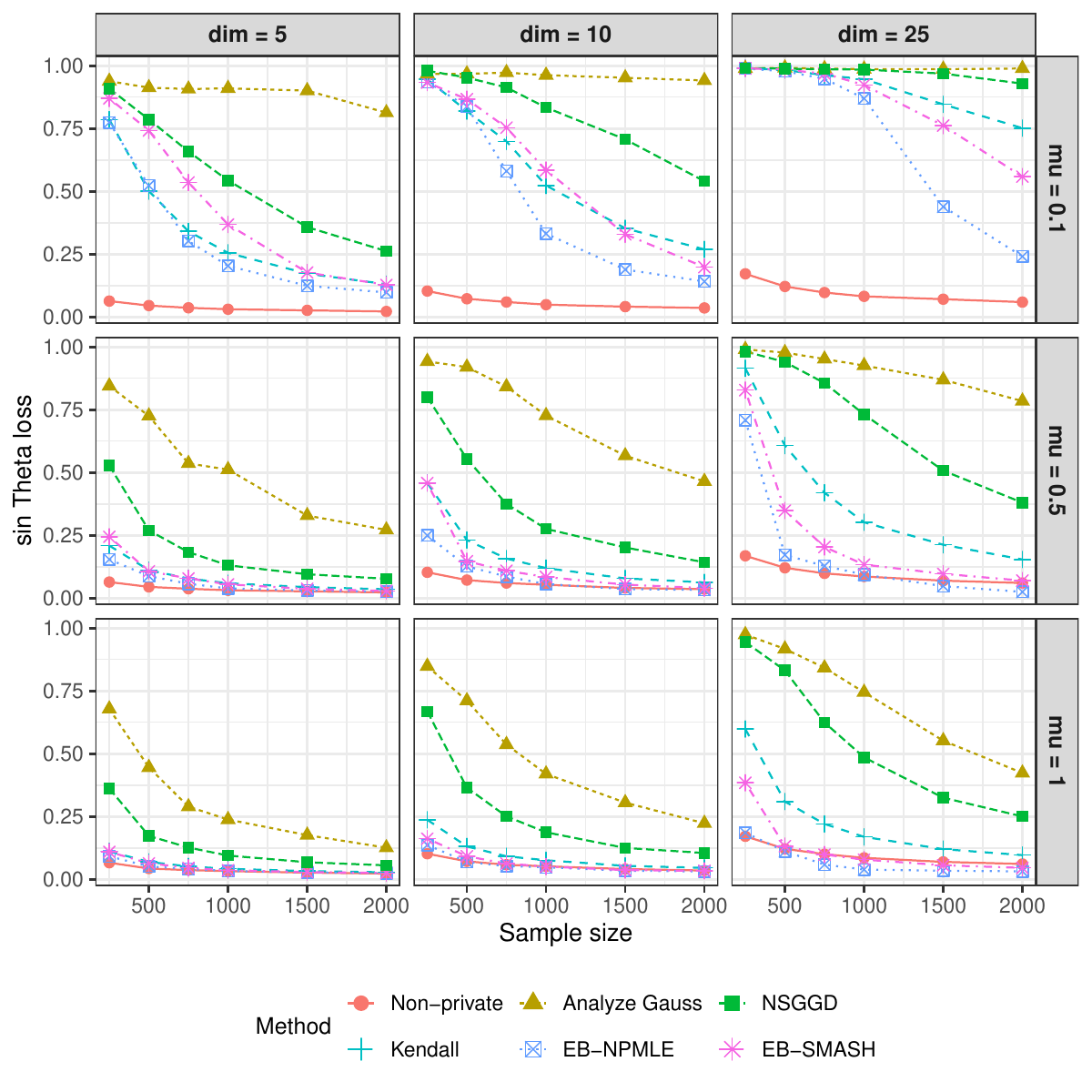}
    \caption{Simulation results for private PCA under a Gaussian distribution.}
    \label{fig:pca-Gaussian}
\end{figure}

\begin{figure}[]
	\centering
	\includegraphics[width=.95\linewidth]{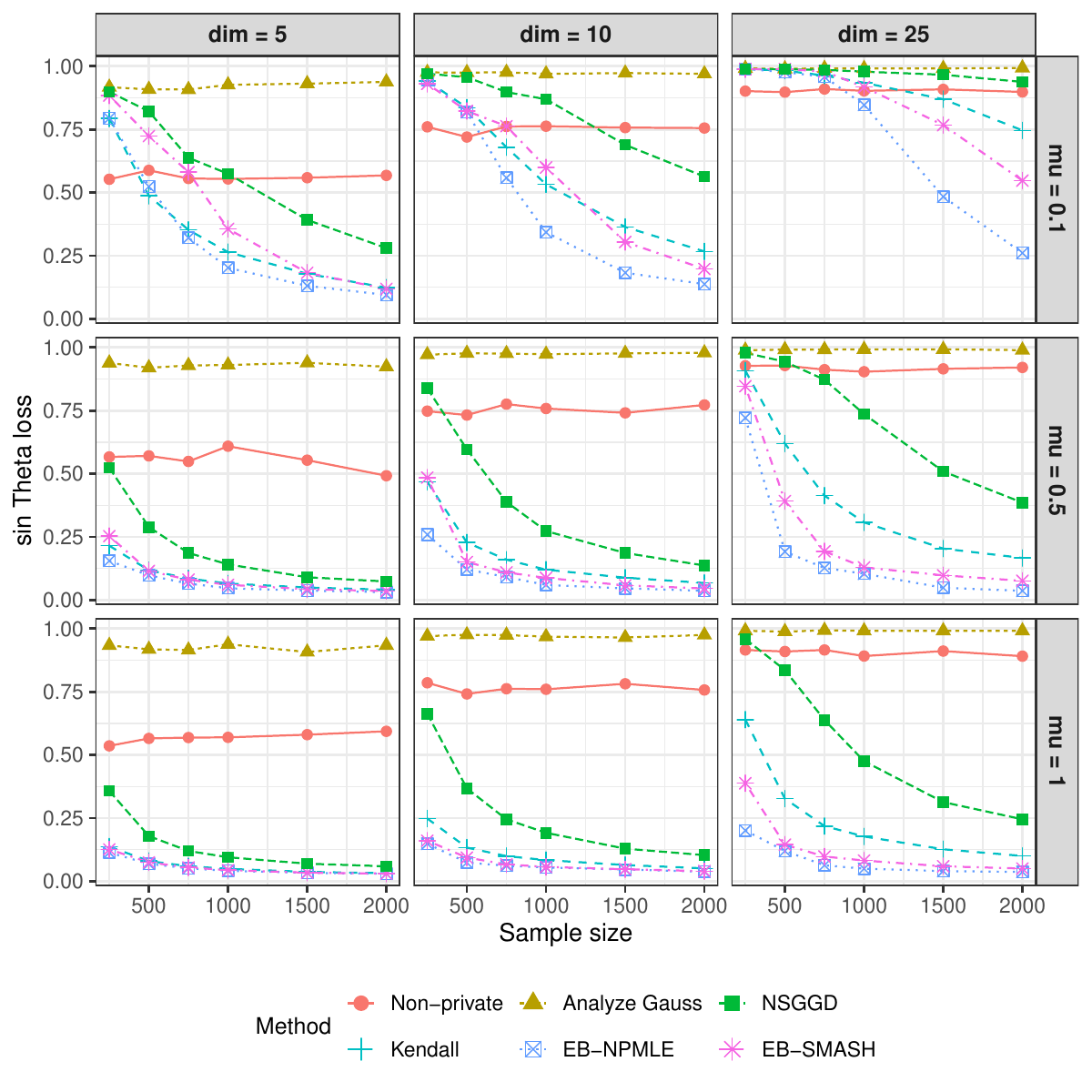}
    \caption{Simulation results for private PCA under a multivariate $t$-distribution.}
    \label{fig:pca-t}
\end{figure}

We conduct numerical simulations to evaluate how the empirical Bayes methods enhance the utility of private PCA. 
We compare the proposed EB methods against the non-private PCA as baseline, Kendall's tau, Analyze Gauss and NSGGD as competing methods.
For the simulation, the similar settings from \citet{kim2025pca} are adopted.
We consider two $d$-dimensional elliptical distributions:
\begin{itemize}
    \item Gaussian: $ X \sim N_d(0, \Sigma) $,
    \item centered multivariate $t$-distribution with degree of freedom 1: $ X \sim t_1(\Sigma) $.
\end{itemize}
For the covariance structure, we consider a two-spiked covariance matrix
\[
\Sigma = (\lambda_1-\lambda_d) v_1 v_1^\top + (\lambda_2-\lambda_d) v_2 v_2^\top + \lambda_d I_d,
\]
where $ (\lambda_1, \lambda_2, \lambda_d) = (10, 5, 1) $ and
\[
v_1 = (1, 1, 1, 1, 0_{d-4})^\top/2 \quad\text{and}\quad v_2 = (1, -1, 1, -1, 0_{d-4})^\top/2. 
\]
For parameters we set $ N \in \{ 250, 500, 750, 1000, 1500, 2000 \} $, $ d \in \{5, 10, 25\} $ and $ \mu \in \{0.1, 0.5, 1\} .$
From each distribution and $(N, d, \mu)$ combination, we sample dataset of size $N$, and applied various $\mu$-GDP mechanisms. 
After obtaining the first two PCs, $ \tilde v_1, \tilde v_2 $, we measure the estimation error using the $ \sin \Theta $ distance \citep{cai2018rateopt} between $ \col([v_1, v_2]) $ and $ \col([\tilde v_1, \tilde v_2]) $, which quantifies the discrepancy between the subspaces spanned by the respective eigenvectors.
%
%
This procedure was repeated 100 times for each setting and the simulation results of averaged losses are reported in Figures \ref{fig:pca-Gaussian} and \ref{fig:pca-t}.

As shown in both figures, our proposed EB-NPMLE consistently achieves the lowest error among all private methods across all simulation settings.
Notably, EB-NPMLE and EB-SMASH substantially improve the original Kendall's tau mechanism.
As expected, the non-private PCA from the sample covariance matrix performs best on Gaussian data (Figure \ref{fig:pca-Gaussian}).
However, as shown in Figure \ref{fig:pca-t}, for heavy-tailed $t_1$ data, the robust methods based on Kendall's tau and our proposed EB methods significantly outperform the standard non-private PCA.
Furthermore, our proposed methods outperform NSGGD, which is also designed to guarantee both privacy and robustness.
Similar to the histogram case in the previous section, the proposed EB methods are most effective in high-dimensional and high-privacy settings, and the performance gaps between EB methods and the other methods become larger in these demanding situations.


\subsection{Private Linear Regression}
For the empirical study of private linear regression, 
we investigate how EB methods enhance the utility of
AdaSSP.
We use several OpenML\footnote{https://www.openml.org} datasets to evaluate the performance of the linear regression methods, where these datasets are frequently used to evaluate private linear regression problems.
The non-private ordinary least squares (OLS) estimator is included as a performance baseline.
For each dataset, we standardize the dataset
and then clip all $X_i$ to have the maximal bound of $\|X_i\|_2 \le 3$
and $Y_i$ to $|Y_i| \le 5$, which ensures that  $\|\Xc\| = 3$ and $\|\Yc\| = 5$.
After the pre-processing, we fit the linear regression estimators
with train-test splitting of 6:4.
The results of averaged test MSE over 100 repetitions are reported in Table \ref{tab:lin-reg}.

Table \ref{tab:lin-reg} clearly show that EB-SMASH, consistently improve regression accuracy over the AdaSSP across most datasets. 
We note that EB-SMASH exhibits consistent improvements, whereas EB-NPMLE is unstable, occasionally yielding significantly worse results than the baseline (e.g., Tasks 361074, 361099).
Remarkably, the test MSE of EB-SMASH is often very close to that of the non-private OLS estimator, indicating that our method effectively denoises the private statistics with minimal loss in utility.

\begin{table}[]
\centering
\caption{Mean and standard error of the test MSE for private linear regression on OpenML datasets, averaged over 100 repetitions.}
\label{tab:lin-reg}
\resizebox{.95\linewidth}{!}{
\begin{tabular}{ccccccc}
\toprule
\multirow[b]{2}{*}{TaskID} & \multirow[b]{2}{*}{$N$} & \multirow[b]{2}{*}{$d$} & \multicolumn{4}{c}{$\mu = .25$}  \\ 
\cmidrule(lr){4-7}
& & & OLS-nonpriv & AdaSSP & EB-NPMLE & EB-SMASH \\ 
\midrule
361072 & 8192  & 22 & 18.968(0.008) & 18.967(0.008) & 18.879(0.010) & \textbf{18.843(0.008)} \\
361073 & 15000 & 27 & 6.488 (0.010) & 6.488 (0.010) & 6.477 (0.010) & \textbf{6.467 (0.010)} \\
361074 & 16599 & 17 & 2.116 (0.002) & 2.116 (0.002) & 10.548(7.359) & \textbf{2.111 (0.001)} \\
361077 & 13750 & 34 & 1.799 (0.002) & 1.799 (0.002) & 2.510 (0.292) & \textbf{1.790 (0.002)} \\
361078 & 20640 &  9 & 21.274(0.002) & 21.274(0.002) & 21.266(0.004) & \textbf{21.255(0.002)} \\
361079 & 22784 & 17 & 16.128(0.002) & 16.128(0.002) & \textbf{16.108(0.003)} & \textbf{16.100(0.002)} \\
361085 & 10081 &  7 & \textbf{0.233 (0.001)} & \textbf{0.233 (0.001)} & 3.572 (3.0566) & \textbf{0.233 (0.001)} \\
361087 & 13932 & 14 & 18.550(0.003) & 18.550(0.003) & 18.530(0.005) & \textbf{18.506(0.002)} \\
361089 & 20640 &  9 & 9.680 (0.005) & 9.680 (0.005) & 9.679 (0.005) & 9.673 (0.005) \\
361092 & 8885  & 83 & 21.710(0.009) & 21.711(0.009) & 21.266(0.026) & \textbf{21.163(0.002)} \\
361093 & 4052  & 13 & 21.069(0.015) & 21.069(0.015) & \textbf{21.016(0.025)} & 20.910(0.014) \\
361094 & 8641  &  6 & \textbf{6.607 (0.009)} & \textbf{6.608 (0.010) }& 7.127 (0.183) & 6.643 (0.010) \\
361098 & 10692 & 18 & 8.882 (0.003) & 8.882 (0.003) & 8.865 (0.005) & \textbf{8.845 (0.002)} \\
361099 & 17379 & 21 & \textbf{1.782 (0.003)} & \textbf{1.783 (0.003)} & 4.297 (1.846) & \textbf{1.777 (0.003)} \\
361102 & 21613 & 20 & 16.998(0.002) & 16.998(0.002) & 16.987(0.004) & \textbf{16.959(0.001)} \\
\bottomrule
\end{tabular}
}
\resizebox{.95\linewidth}{!}{
\begin{tabular}{ccccccc}
\toprule
\multirow[b]{2}{*}{TaskID} & \multirow[b]{2}{*}{$N$} & \multirow[b]{2}{*}{$d$} &  \multicolumn{4}{c}{$\mu = 1$} \\ 
\cmidrule(lr){4-7}
& & & OLS-nonpriv & AdaSSP & EB-NPMLE & EB-SMASH \\ 
\midrule
361072 & 8192  & 22 & 18.981(0.008) & 18.982(0.008) & 18.934(0.008) & \textbf{18.890(0.008)}\\ 
361073 & 15000 & 27 & 6.441 (0.011) & 6.441 (0.011) & 6.436 (0.012) & 6.430 (0.011)\\ 
361074 & 16599 & 17 & \textbf{2.110 (0.002)} & \textbf{2.110 (0.002)} & 2.111 (0.002) & \textbf{2.106 (0.002)}\\ 
361077 & 13750 & 34 & 1.801 (0.002) & 1.801 (0.002) & 1.798 (0.002) & \textbf{1.790 (0.002)}\\ 
361078 & 20640 &  9 & 21.273(0.002) & 21.273(0.002) & 21.270(0.002) & \textbf{21.264(0.002)}\\ 
361079 & 22784 & 17 & 16.121(0.002) & 16.121(0.002) & 16.117(0.002) & \textbf{16.113(0.002)}\\ 
361085 & 10081 &  7 & \textbf{0.233 (0.001)} & \textbf{0.233 (0.001)} & 0.308 (0.049) & \textbf{0.233 (0.001)}\\ 
361087 & 13932 & 14 & 18.551(0.003) & 18.551(0.003) & 18.539(0.003) & \textbf{18.524(0.003)}\\ 
361089 & 20640 &  9 & 9.695 (0.006) & 9.695 (0.006) & 9.695 (0.006) & 9.692 (0.006)\\ 
361092 & 8885  & 83 & 21.712(0.009) & 21.713(0.009) & 21.342(0.009) & \textbf{21.176(0.003)}\\ 
361093 & 4052  & 13 & 21.051(0.018) & 21.051(0.018) & 21.019(0.018) & \textbf{20.981(0.019)}\\ 
361094 & 8641  &  6 & \textbf{6.567 (0.009)} & \textbf{6.564 (0.008)} & 9.806 (2.271) & 6.616 (0.009)\\ 
361098 & 10692 & 18 & 8.878 (0.002) & 8.878 (0.002) & 8.866 (0.002) & \textbf{8.853 (0.002)}\\ 
361099 & 17379 & 21 & \textbf{1.789 (0.003) }& \textbf{1.789 (0.003)} & 8.101 (6.269) & \textbf{1.784 (0.003)}\\ 
361102 & 21613 & 20 & 16.994(0.002) & 16.994(0.002) & 16.988(0.002) & \textbf{16.980(0.002)}\\ 
\bottomrule
\end{tabular}
}
\end{table}

\section{Conclusion}
We demonstrated that empirical Bayes estimators can substantially boost the utility of the simple Gaussian additive mechanism across a range of statistical estimation tasks, including histograms, PCA, and linear regression.
This denoising process benefits from the post-processing property of differential privacy, meaning it does not compromise privacy.
Importantly, the process is entirely adaptive and automated, eliminating the need for elaborate private algorithm design or delicate parameter tuning.
The promising results suggest that extensions to other forms of data release or statistical analysis, such as density and function estimation, or individual treatment effect estimation in causal inference, are both feasible and compelling. Thus, our contribution lays the groundwork for a general and influential framework that can significantly improve the practical utility of differentially private methods across diverse domains.

While we mainly consider Gaussian differential privacy, our EB approach can also be adapted to other notions of privacy, such as $ (\varepsilon, \delta) $-DP and $\rho$-zCDP \citep{bun2016concentrated}, with the Gaussian additive mechanism.
One important remark is that EB-NPMLE is not limited only to the Gaussian mechanism, and can be applied to any form of additive mechanism where the density function of the additive noise is available. 
For instance, by applying EB-NPMLE to the Laplace mechanism, it is possible to obtain the $\varepsilon$-DP mechanism with improved performance compared to the original Laplace mechanism.
Furthermore, as a denoising process, EB approach can applied to any output of differentially private mechanisms.

We leave a theoretical analysis of the empirical successes, as well as extensions to other statistical tasks and privacy mechanisms, as important directions for future work.

\acks{
This work was supported by the National Research Foundation of Korea (NRF) grant funded by the Korea government (MSIT) (RS-2023-00218231, RS-2024-00333399, RS-2025-00556575).
}



\appendix

\section{EM algorithm for NPMLE}
Here, we explain how the discretized NPMLE problem of (5) in Section 3.1 of the main paper, can be solved using the EM algorithm.
We follow Section 2.3 of \citet{jiang2009general}.

Let $ \min Y_i =  a_1 < a_2 < \dots < a_m = \max Y_i $ be equally spaced grid points for the support, 
and consider
\begin{equation*}
    \Gc = \left\{G = \sum_{j=1}^m w_j \delta_{a_j} ~\bigg|~ \sum_{j=1}^m w_j = 1,~ w_j \ge 0 \right\},
\end{equation*}
where $ \delta_{a_j} $ is a point mass at $ a_j $.
For $G = \sum_{j=1}^m w_j \delta_{a_j} \in \Gc $, it holds that
\begin{equation*}
    f_G(x) = \int p(x \;|\; \theta) dG(\theta) = \sum_{j=1}^m w_j p(x \;|\; a_j).
\end{equation*}
Hence, corresponding optimization problem of (5) becomes
\[
    \max_{w_1, \dots, w_m \ge 0} ~
    \prod_{i=1}^n \left(\sum_{j=1}^m w_j p(Y_i \;|\; a_j) \right)
    ~
    \mbox{ s.t. } ~ \sum_{j=1}^{m} w_j = 1,
\]
or equivalently,
\[
    \max_{w_1, \dots, w_m \ge 0} ~
    \sum_{i=1}^n \log\left(\sum_{j=1}^m w_j p(Y_i \;|\; a_j) \right)
    ~
    \mbox{ s.t. } ~ \sum_{j=1}^{m} w_j = 1.
\]
The corresponding EM algorithm updates the weights $ \{w_j\} $ as follows:
\begin{equation} \label{eq:supp-EM}
    \widehat w_j^{(k)} 
    = \frac{1}{n} \sum_{i=1}^n \frac{\widehat w_j^{(k-1)}p(Y_i | a_j)}{\sum_{\ell = 1}^m \widehat w_{\ell}^{(k-1)}p(Y_i | a_{\ell})}.
\end{equation}
Although several initialization options exist, we set $\widehat w_j^{(0)} = 1/m$ for all $j$.
For a more detailed discussion, we refer the reader to \citet{jiang2009general}.

Once the weights $\{\widehat a_j\}$ are obtained, the corresponding NPMLE-EB estimator introduced in Section 3.1 becomes
\begin{equation} \label{eq:supp-npmle-eb}
    \hat \theta_i 
    = \frac{\int \theta p(Y_i |\theta) d\widehat G(\theta)}{\int p(Y_i | \theta)d\widehat G(\theta)}
    = \frac{\sum_{j=1}^m  \widehat{w}_j a_j p(Y_i | a_j) }{\sum_{j=1}^m  \widehat{w}_j p(Y_i | a_j)}.
\end{equation}

\section{EB-NPMLE for Laplace mechanism} \label{appendix:Lap}
One of the most popular notions of DP is $\varepsilon$-DP, proposed by \cite{dwork2006calibrating}.

\begin{definition}
    Let $\varepsilon > 0$.
    Then a randomized mechanism $ \Mc $ satisfies $\varepsilon$-differential privacy (DP)
    if for any measurable subset $E$ and for any neighboring datasets  $S \sim S'$,
    \[
        \Pb(\Mc(S) \in E) \le e^{\varepsilon} \Pb(\Mc(S') \in E).
    \]
\end{definition}

Let $S = (x_1, \dots, x_n) \in \Xc^n$ be a dataset and 
consider a statistic $ \theta(S) \in \Rb^p $.
The $\ell_1$-sensitivity of $\theta$ is defined as $\Delta_1 = \sup_{S \sim S'} \|\theta(S) - \theta(S')\|_1.$ 
\textit{Laplace mechanism} adds a Laplace noise calibrated with $\Delta_1$ to $\theta(S)$ to satisfy $\varepsilon$-DP.

\begin{proposition}
    Let Laplace additive mechanism
    \[
        \Mc_{Lap}(S; b) = \theta(S) + \xi,
    \]
    where 
    $ \xi_i \overset{i.i.d.}{\sim} \mbox{Lap}(0; b) $ for $i = 1, \dots, p$.
    Then $ \Mc_{Lap} $ satisfies $\varepsilon$-DP if $b \ge \Delta_1 / \varepsilon$.
\end{proposition}

Set $b = \Delta_1 / \varepsilon$, and denote 
$\Mc_{Lap}(S) = (M_1, \dots, M_p)$ and $\theta(S) = (\theta_1, \dots, \theta_p)$.
As in the Gaussian case, the Laplace mechanism follows the conditional distributional relationship as
\[
    M_i \:|\: \theta_i \overset{ind.}{\sim} \mbox{Lap}(\theta_i; b),
\]
where $ \mbox{Lap}(\theta_i; b) $ has the density function 
\begin{equation} \label{eq:supp-lap}
    p_{Lap}(x \;|\; \theta_i) = \frac{\varepsilon}{2\Delta_1}\exp\left(\frac{-\varepsilon|x-\theta_i|}{\Delta_1}\right).
\end{equation}
Then, EB-NPMLE can be applied to the output of $\Mc_{Lap}$ as a post-processing 
by solving \eqref{eq:supp-EM} and obtaining the estimator \eqref{eq:supp-npmle-eb} with the density \eqref{eq:supp-lap}.
We denote this mechanism as the \textit{Lap-NPMLE} mechanism.

\subsection{Simulation study}
Here, we conduct a simulation study on releasing histograms, following the setup in Section 6.1 of the main paper.
Observe that $\Delta_1(\theta^{hist}) = 2$.
So, the Laplace mechanism with the scale parameter $b = 2 / \varepsilon$ satisfies 
$\varepsilon$-DP.
We compare the Laplace mechanism, the Lap-NPMLE mechanism, and the non-private histogram, which serves as a baseline.
To fit Lap-NPMLE, we set $m = 1000$ and iterate the EM algorithm \eqref{eq:supp-EM} for 100 times, i.e., we run the algorithm until $k = 100$.
For the data generation model, we use the same settings as in Section 6.1.
The simulations are conducted in a manner similar to that of Section 6.1, except for the privacy parameters.

\begin{figure}[]
	\centering
    \includegraphics[width=.68\linewidth]{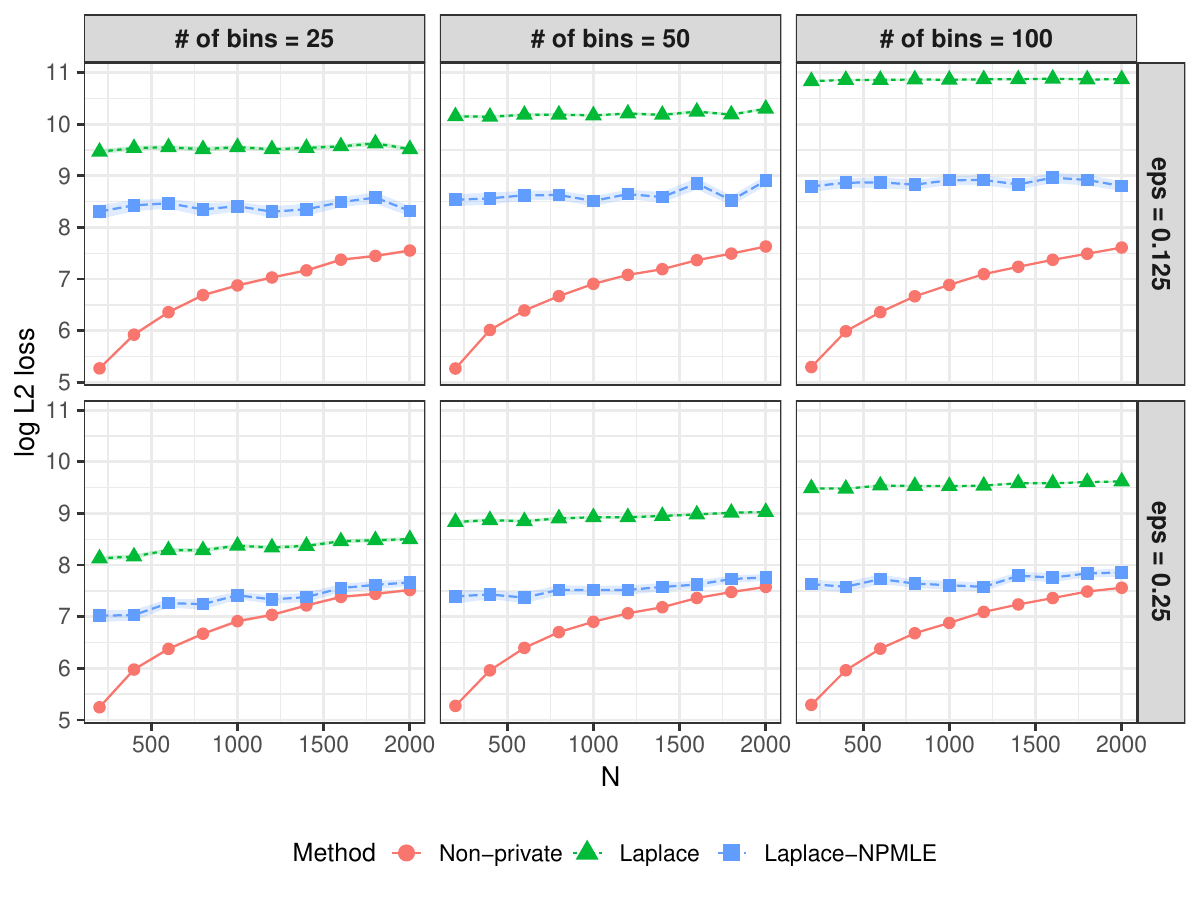}
    \caption{Average population $L_2$-costs of private histogram release under Model I with high-privacy setting.}
    \label{fig:hist-lap-M1-highpriv}
\end{figure}

\begin{figure}[]
	\centering
    \includegraphics[width=.68\linewidth]{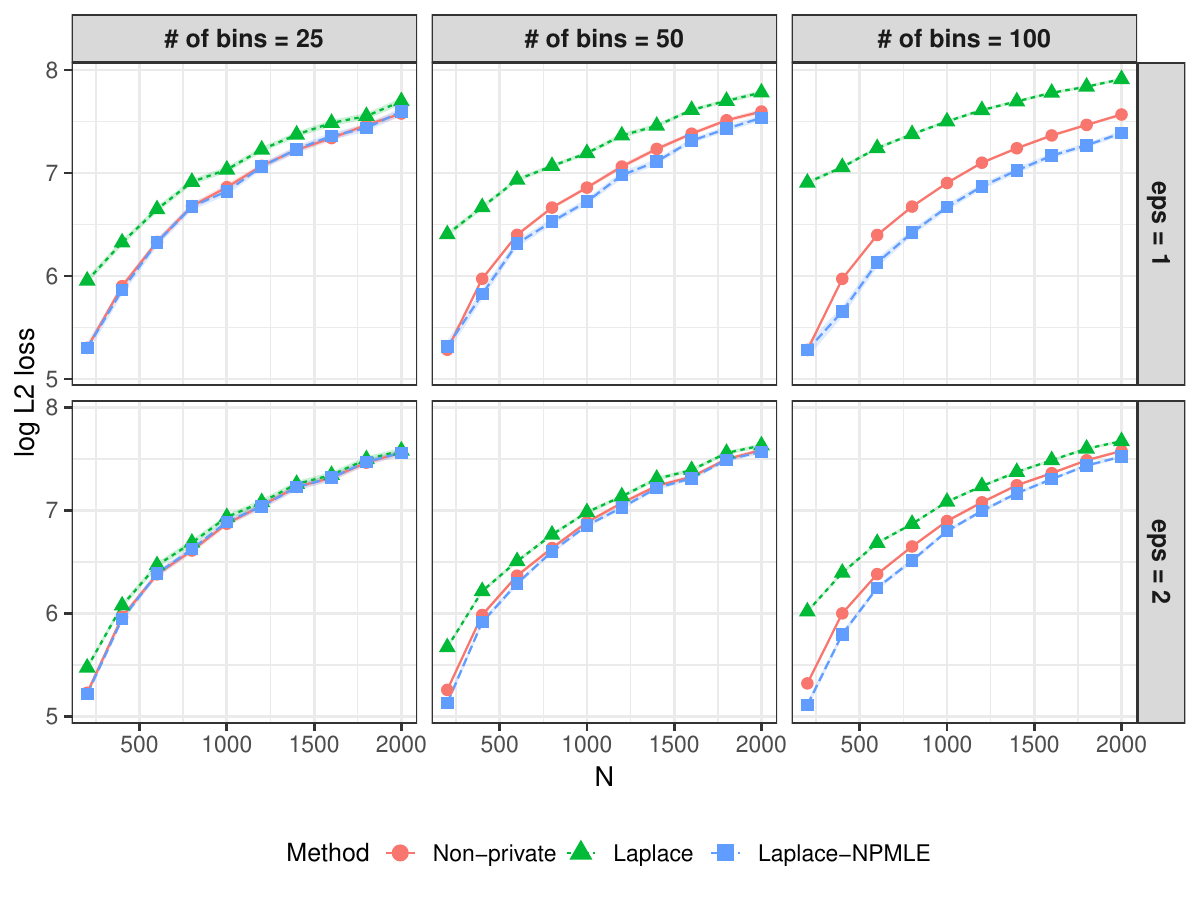}
    \caption{Average population $L_2$-costs of private histogram release under Model I with low-privacy setting.}
    \label{fig:hist-lap-M1-lowpriv}
\end{figure}

\begin{figure}[]
	\centering
    \includegraphics[width=.68\linewidth]{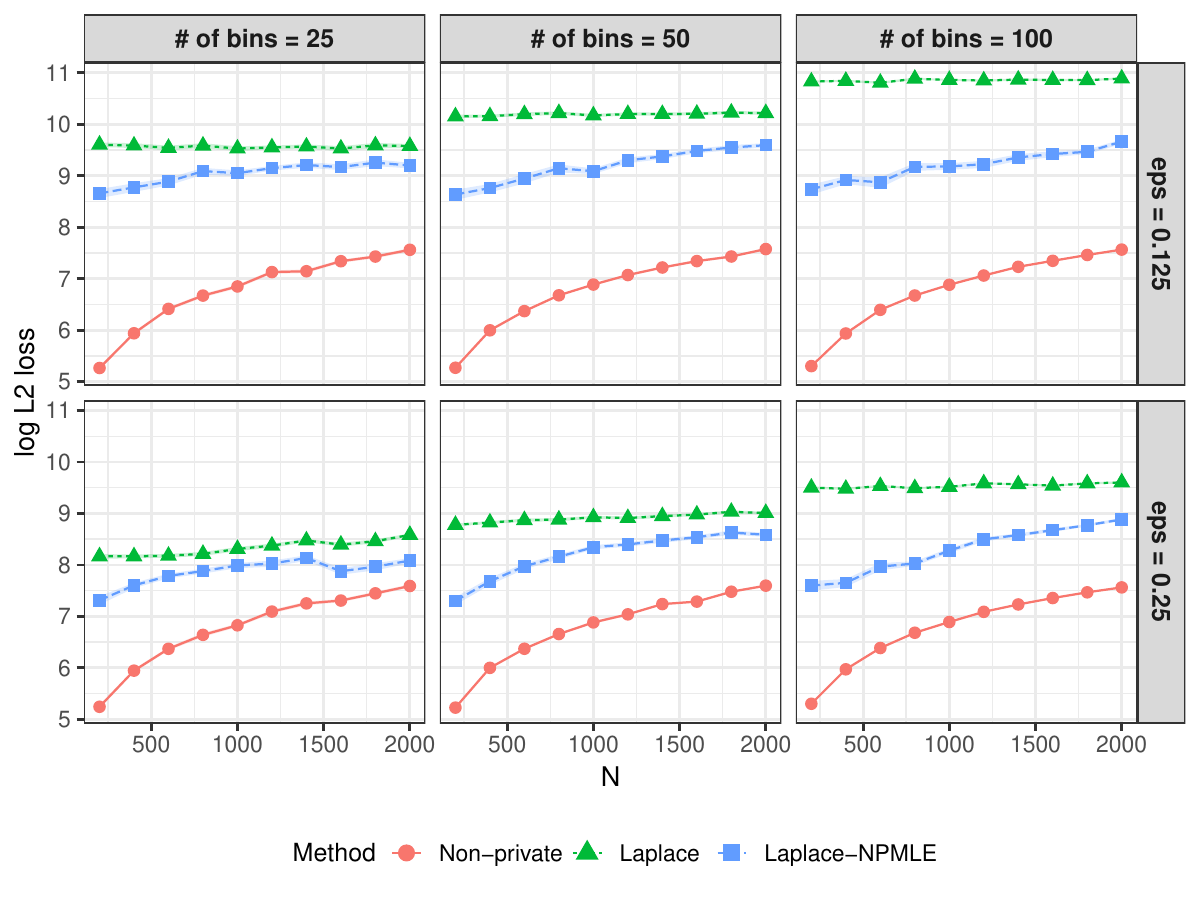}
    \caption{Average population $L_2$-costs of private histogram release under Model II with high-privacy setting.}
    \label{fig:hist-lap-M2-highpriv}
\end{figure}

\begin{figure}[]
	\centering
    \includegraphics[width=.68\linewidth]{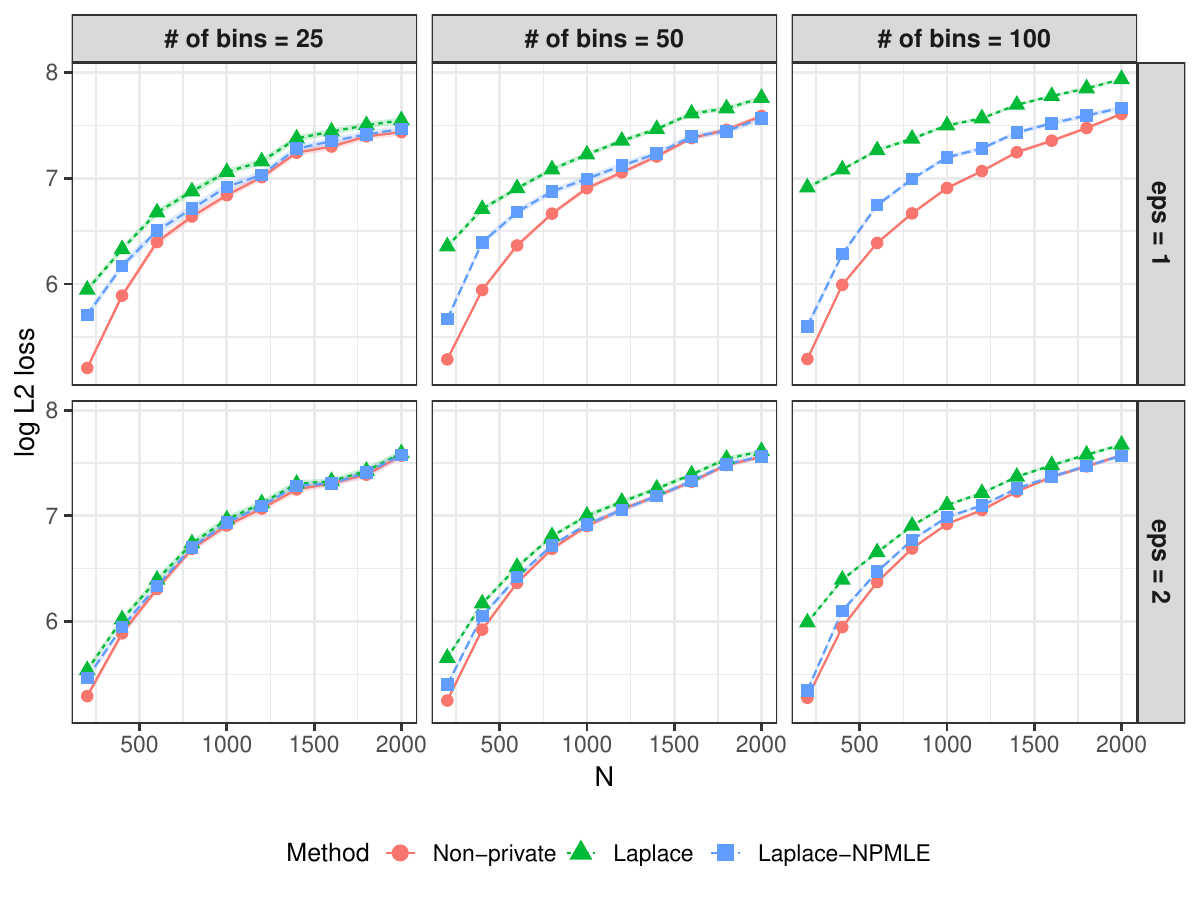}
    \caption{Average population $L_2$-costs of private histogram release under Model II with low-privacy setting.}
    \label{fig:hist-lap-M2-lowpriv}
\end{figure}

The results are described in Figures \ref{fig:hist-lap-M1-highpriv} to \ref{fig:hist-lap-M2-lowpriv}.
In all cases, Lap-NPMLE outperforms the original Laplace mechanism.
Similar to the Gaussian cases, the performance gap between the Laplace mechanism and Lap-NPMLE 
widens as the privacy guarantee becomes stronger (i.e., for smaller $\varepsilon$) 
and as the dimension $p$ increases.
In most cases, the non-private histogram outperforms Lap-NPMLE, except for the 
Model I with low privacy and high-dimensional case; see Figure \ref{fig:hist-lap-M1-lowpriv}.
This is in contrast to the EB-Gaussian mechanisms, where both EB-NPMLE and EB-SMASH outperform
the non-private histogram, as we have observed in Section 6.1 of the main paper.

\newpage
\bibliography{jmlr-bib}

\end{document}